\documentclass{article}

\PassOptionsToPackage{numbers, compress}{natbib}

\usepackage[preprint]{neurips_2025}

\usepackage[utf8]{inputenc} 
\usepackage[T1]{fontenc}    
\usepackage{hyperref}       
\usepackage{url}            
\usepackage{booktabs}       
\usepackage{amsfonts}       
\usepackage{nicefrac}       
\usepackage{microtype}      
\usepackage{xcolor}         

\usepackage{amsmath}
\usepackage{mathrsfs}
\usepackage{amssymb} 
\usepackage[linesnumbered,ruled,vlined]{algorithm2e}
\usepackage{xcolor}
\usepackage{listings}
\usepackage{tcolorbox}
\usepackage{enumitem}
\usepackage{dsfont}
\usepackage{multirow}
\usepackage{makecell}
\usepackage{threeparttable}  
\usepackage{arydshln} 
\usepackage{float}  
\usepackage{caption}

\newcommand{\thinkstart}{\textcolor[HTML]{2e57d0}{<think>}}
\newcommand{\thinkend}{\textcolor[HTML]{2e57d0}{</think>}}

\newcommand{\sqlstart}{\textcolor[HTML]{ffb713}{{<intermediate\_sql>}}}
\newcommand{\sqlend}{\textcolor[HTML]{ffb713}{{</intermediate\_sql>}}}

\newcommand{\resultstart}{\textcolor[HTML]{16b69e}{{<result>}}}
\newcommand{\resultend}{\textcolor[HTML]{16b69e}{{</result>}}}

\newcommand{\finalstart}{\textcolor[HTML]{bf0040}{{<final\_sql>}}}
\newcommand{\finalend}{\textcolor[HTML]{bf0040}{{</final\_sql>}}}

\title{ReEx-SQL: Reasoning with Execution-Aware Reinforcement Learning for Text-to-SQL}

\author{%
    Yaxun~Dai\textsuperscript{\rm 1}\footnotemark[1],~Wenxuan~Xie\textsuperscript{\rm 3},~Xialie~Zhuang\textsuperscript{\rm 4},~Tianyu~Yang\textsuperscript{\rm 5},~Yiying~Yang\textsuperscript{\rm 2},\\ 
\textbf{Haiqin~Yang\textsuperscript{\rm 6},~Yuhang~Zhao\textsuperscript{\rm 2},~Pingfu~Chao\textsuperscript{\rm 1}\footnotemark[2],~Wenhao~Jiang\textsuperscript{\rm 2}\footnotemark[2]} \\
  \textsuperscript{\rm 1}Soochow University\\
  \textsuperscript{\rm 2}Guangdong Laboratory of Artificial Intelligence and Digital Economy (SZ) \\
  \textsuperscript{\rm 3}South China University of Technology \\
  \textsuperscript{\rm 4}University of Chinese Academy of Sciences \\
  \textsuperscript{\rm 5}Alibaba DAMO Academy \\
  \textsuperscript{\rm 6}International Digital Economy Academy (IDEA), China \\
\texttt{yaxundai@gmail.com},~\texttt{pfchao@suda.edu.cn},~\texttt{cswhjiang@gmail.com}
}

\begin{document}

\maketitle
{
\renewcommand{\thefootnote}{\fnsymbol{footnote}}
\footnotetext[1]{Work done during an internship at the Guangdong Laboratory of Artificial Intelligence and Digital Economy (SZ).}
\footnotetext[2]{Corresponding authors.}
}

\begin{abstract}
In Text-to-SQL, execution feedback is essential for guiding large language models (LLMs) to reason accurately and generate reliable SQL queries.  However, existing methods treat execution feedback solely as a post-hoc signal for correction or selection, failing to integrate it into the generation process.  This limitation hinders their ability to address reasoning errors as they occur, ultimately reducing query accuracy and robustness.  To address this issue, we propose \textbf{ReEx-SQL} (\textit{Reasoning with Execution-Aware Reinforcement Learning}), a framework for Text-to-SQL that enables models to interact with the database during decoding and dynamically adjust their reasoning based on execution feedback.  ReEx-SQL introduces an execution-aware reasoning paradigm that interleaves intermediate SQL execution into reasoning paths, facilitating context-sensitive revisions.  It achieves this through structured prompts with markup tags and a stepwise rollout strategy that integrates execution feedback into each stage of generation.  To supervise policy learning, we develop a composite reward function that includes an exploration reward, explicitly encouraging effective database interaction.  Additionally, ReEx-SQL adopts a tree-based decoding strategy to support exploratory reasoning, enabling dynamic expansion of alternative reasoning paths. Notably, ReEx-SQL achieves 88.8\% on Spider and 64.9\% on BIRD at the 7B scale, surpassing the standard reasoning baseline by 2.7\% and 2.6\%, respectively. It also shows robustness, achieving 85.2\% on Spider-Realistic with leading performance. In addition, its tree-structured decoding improves efficiency and performance over linear decoding, reducing inference time by 51.9\% on the BIRD development set.
\end{abstract}

\begin{figure}[t]
    \includegraphics[width=1\columnwidth]{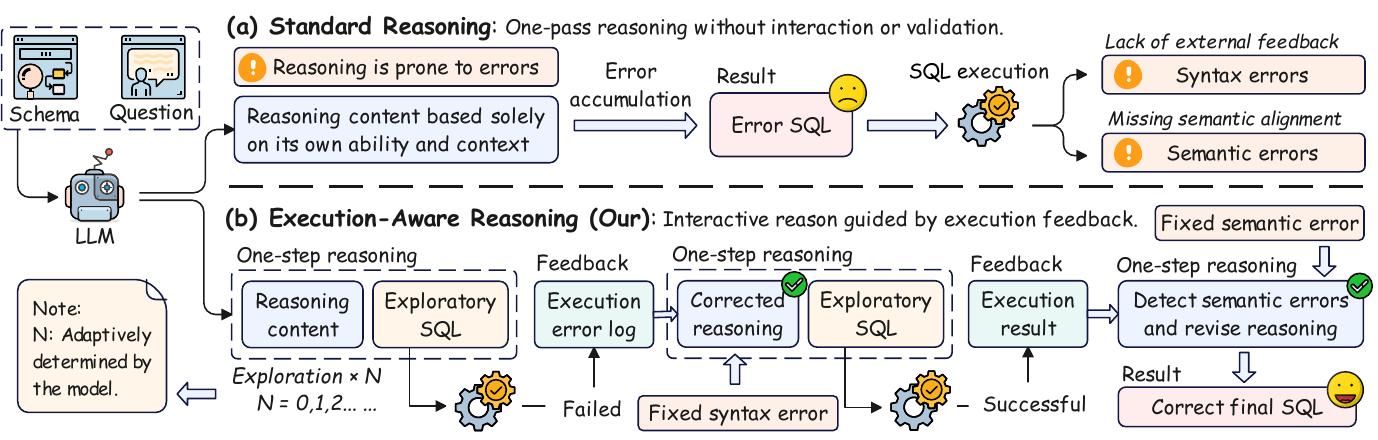}
    \caption{Standard reasoning vs. Execution-aware reasoning.  Standard reasoning (e.g., Chain-of-Thought~\citep{DBLP:conf/nips/cot}) generates a reasoning path without validation, making it prone to error accumulation, syntax errors, and semantic misalignment. In contrast, execution-aware reasoning validates each step via execution feedback, enabling the detection and correction of both syntactic and semantic errors.}
    \label{fig:motivation}
\end{figure}

\section{Introduction}

The Text-to-SQL task aims to map natural language questions to executable SQL queries over a given database and serves as a key technique for enabling natural language interfaces to structured data~\citep{DBLP:conf/coling/Deng0022,DBLP:journals/vldb/KatsogiannisMeimarakisK23,DBLP:journals/corr/surveynl2sql}.  It plays a critical role in the building intelligent agents, requiring precise reasoning over user intent, database schema, and SQL logic~\citep{DBLP:journals/TKDE/ZhangWSWTQCWY24,DBLP:journals/corr/liu2024survey,DBLP:journals/corr/surveynl2sql}.  In recent years, large language models (LLMs) have been widely adopted in Text-to-SQL  task due to their strong reasoning capabilities~\citep{Alpha-SQL,CHESS,DBLP:conf/acl/RAT-SQL,DBLP:conf/sigmod/Finsql,xiyansql,supersql,PURPLE}.  In particular, Chain-of-Thought (CoT)~\citep{DBLP:conf/nips/cot}, which introduces intermediate reasoning steps prior to generating the final SQL query, has proven effective in improving accuracy and robustness when handling complex queries~\citep{DBLP:conf/nips/din_sql,CHESS,cot_survey,cot_act_sql,ExCoT}.  However, as illustrated in Figure~\ref{fig:motivation}, CoT-based standard reasoning still exhibit significant limitations.  Specifically, LLMs relies solely on static context and its own reasoning capacity during generation, lacking mechanisms to verify or revise intermediate decisions.  This often results in: (1) error accumulation from incorrect reasoning steps that the model cannot self-correct~\citep{DBLP:conf/acl/multask,DBLP:renda_dpo_sql}; (2) syntactic errors due to the absence of execution validation~\citep{ExCoT,NL2SQL-BUGs,SQLCritic}; and (3) semantic misalignment, where the generated SQL fails to faithfully reflect the user’s intent~\citep{sql_error_survey,reasoning_errors,Multi_grained_Error}.

Recent studies have mainly pursued two approaches to mitigate reasoning errors~\citep{CHESS,DBLP:journals/corr/supersql,DBLP:conf/nips/din_sql,pourreza2024chase}. The first focuses on improving the CoT reasoning process itself, often via reinforcement learning, to encourage more accurate intermediate steps~\citep{Alpha-SQL,SQL-o1,EllieSQL,DBLP:renda_dpo_sql,cot_act_sql,ExCoT}. However, it still fails to address the core issue—the lack of execution feedback—leaving the model unable to validate its reasoning during generation.  The second treats verification as a separate post-decoding task, intended to detect and correct syntactic or semantic errors in the generated SQL~\citep{pourreza2024chase,DBLP:journals/corr/macsql,DBLP:conf/nips/din_sql,ROUTE}. While this can improve overall correctness, decoupling verification from generation risks altering otherwise correct predictions, potentially introducing new errors instead of preventing them~\citep{NL2SQL-BUGs,sql_error_survey}.

To address these challenges, we propose \textbf{ReEx-SQL} (\textit{Reasoning with Execution-Aware Reinforcement Learning}), a novel framework that incorporates execution feedback into the decoding process, enabling the model to interact with the database and iteratively refine its reasoning.  Specifically, ReEx-SQL introduces an execution-aware reasoning paradigm, where exploratory SQLs are interleaved with their execution results to guide the reasoning process until the final SQL is produced.  This is achieved through a structured prompting format with explicit markup tokens—such as \textcolor[HTML]{2e57d0}{\texttt{<think>}}, \textcolor[HTML]{ffb713}{\texttt{<intermediate\_sql>}}, \textcolor[HTML]{16b69e}{\texttt{<result>}}, and \textcolor[HTML]{bf0040}{\texttt{<final\_sql>}}—to clearly delineate reasoning steps, exploratory queries, feedback, and final outputs.   To enhance this interactive reasoning process, ReEx-SQL employs reinforcement learning with a composite reward function. We adopt the Grouped Reinforcement Policy Optimization (GRPO)~\citep{GRPO} algorithm and extend it with a stepwise, execution-aware rollout mechanism to enable dynamic feedback during training.  This process is guided by a composite reward function comprising five components—format, execution, exact match, entity match, and exploration—each providing complementary supervision to guide the model toward generating correct, executable, and semantically faithful SQL queries.  To better explore the reasoning space, ReEx-SQL uses a tree-structured decoding strategy, where each node represents an interaction step. This structure supports adaptive path expansion and significantly improves decoding efficiency and flexibility compared to linear decoding.  Our contributions can be summarized as follows:
\begin{itemize}[leftmargin=1em]
  \item We propose ReEx-SQL, a framework that incorporates execution feedback to iteratively refine reasoning and adopts tree-structured decoding for broader reasoning and faster inference.
  
  \item We extend GRPO with execution-aware rollouts for dynamic interaction during training, and design a composite reward function that jointly optimizes SQL format, executability, and semantic fidelity.

  \item ReEx-SQL-7B achieves strong execution accuracy across both in-domain (88.8\% on Spider, 64.9\% on BIRD) and cross-domain (78.6\% on Spider-Syn, 85.2\% on Spider-Realistic), outperforming prior work.  Compared to standard reasoning, it improves the EX score by 2.6\% and reduces syntax errors from 4.5\% to 1.9\% on the BIRD development set.  Its tree-structured decoding boosts both accuracy and efficiency, cutting per-sample inference time by 51.9\% compared to linear decoding.

\end{itemize}

\section{Related Work}
\subsection{Reasoning Strategies in Text-to-SQL}

Text-to-SQL aims to bridge the gap between natural language questions and structured databases by translating user inputs into executable SQL queries~\citep{DBLP:journals/corr/liu2024survey,DBLP:journals/TKDE/ZhangWSWTQCWY24,DBLP:journals/corr/surveynl2sql,DBLP:journals/corr/supersql}. With the rapid progress of pre-trained language models, recent research has focused on enhancing the reasoning capabilities required to generate complex SQL queries~\citep{CHESS,DBLP:journals/corr/supersql,DBLP:conf/nips/din_sql,pourreza2024chase}. Existing approaches mainly fall into two categories, each aiming to improve model reasoning performance.  The first category aims to strengthen the reasoning process itself, primarily by optimizing Chain-of-Thought (CoT) reasoning~\citep{Alpha-SQL,SQL-o1,EllieSQL,DBLP:renda_dpo_sql,cot_act_sql,ExCoT}. \citet{DBLP:renda_dpo_sql} argue that jointly supervising both intermediate reasoning and final SQL may dilute training objectives. To address this, recent studies like ExCoT~\citep{ExCoT} and EllieSQL~\citep{EllieSQL} adopt reinforcement learning techniques, such as Direct Preference Optimization (DPO)~\citep{DPO}, to build preference-based training signals. While reinforcement learning improves reasoning quality, it still lacks execution feedback, limiting its ability to produce SQL queries that are fully executable and aligned with the database content.  The second category simplifies the task by decomposing it into two stages: initial SQL generation followed by SQL correction. For example, CHASE-SQL~\citep{pourreza2024chase}, MAC-SQL~\citep{DBLP:journals/corr/macsql}, DIN-SQL~\citep{DBLP:conf/nips/din_sql} and MAG-SQL~\citep{xie2024magsqlmultiagentgenerativeapproach} introduce agent-based correction, while ROUTE-SQL~\citep{ROUTE} treats SQL correction as an auxiliary task within a multi-task learning framework. Although this approach reduces reasoning complexity, prior work~\citep{cot_act_sql,NL2SQL-BUGs,sql_error_survey} shows that post-hoc correction may overwrite correct SQL and introduce new errors, disrupting the overall reasoning process.

\subsection{Execution Feedback in Text-to-SQL}
Execution feedback serves as a critical external signal to improve the reasoning capabilities of large language models. \citet{reasoning_errors} demonstrate that even imperfect feedback can significantly enhance self-correction abilities. In Text-to-SQL task, execution feedback operates at two primary levels. First, by leveraging execution error signals, it aids in correcting syntactic errors in generated SQL queries, thereby reducing the likelihood of execution failures~\citep{pourreza2024chase,DBLP:journals/corr/macsql,Alpha-SQL,sql_error_survey}. Second, by utilizing execution results, it evaluates whether the generated queries align with the user's intent, thereby mitigating semantic errors to some extent~\citep{sql_error_survey,DBLP:journals/pvldb/dail,DBLP:journals/corr/c3}. Additionally, some works employ execution-based self-consistency strategies to select the most appropriate SQL query from a pool of candidates, thereby enhancing the overall quality of results~\citep{DBLP:journals/pacmmod/codes,supersql,DBLP:renda_dpo_sql,DBLP:conf/aaai/resdsql,CHESS}.  Inspired by these findings, we propose a novel approach that treats execution feedback not merely as a post hoc correction signal but as a dynamic, environment-grounded mechanism that guides the model’s exploration of the database structure throughout the reasoning process.

\section{Methodology}

\subsection{Execution-Aware Reinforcement Learning}

Reinforcement learning has recently demonstrated its effectiveness in enhancing the reasoning capabilities of LLMs without relying on supervised data~\citep{Deepseek-r1,DBLP:renda_dpo_sql,DeepSeek-Coder}. Inspired by related work~\citep{Search-R1,ReSearch,R1_Searcher}, we apply reinforcement learning to enable autonomous SQL generation without manually limiting database interaction frequency. Specifically, we adopt the GRPO algorithm~\citep{GRPO}, which estimates baselines through group-wise relative rewards, eliminating the need for a separate value model. This reduces memory and computation costs while maintaining strong performance.

Formally, given an existing policy $\pi_{\theta_{\text{old}}}$ and a reference policy $\pi_{\theta_{\text{ref}}}$, for each input $x$ sampled from the data distribution $\mathcal{D}$, we generate $G$ rollouts $\tau = \{ y_i \}_{i=1}^{G}$, where each $y_i$ is drawn from $\pi_{\theta_{\text{old}}}(\cdot|x; \mathcal{E})$. Here, $\mathcal{E}$ denotes the SQL executor, and each rollout $y_i$ can be interpreted as $\pi_{\theta_{\text{old}}}(\cdot|x) \otimes \mathcal{E}$, where $\otimes$ represents the interleaved process of execution and reasoning, detailed explanations are provided in Section~\ref{section_rollout}. The objective of GRPO is to update the policy $\pi_{\theta}$ by maximizing the following objective:
{\small
\begin{multline}
\mathcal{J}(\theta) = \mathbb{E}_{x \sim \mathcal{D}, \{y_i\}_{i=1}^G \sim \pi_{\theta_{\text{old}}}(\cdot|x;\mathcal{E})} \\
\left[
\frac{1}{G} \sum_{i=1}^G \left(
\min\left( \frac{\pi_\theta(y_i|x;\mathcal{E})}{\pi_{\theta_{\text{old}}}(y_i|x;\mathcal{E})} A_i, \, \text{clip}\left( \frac{\pi_\theta(y_i|x;\mathcal{E})}{\pi_{\theta_{\text{old}}}(y_i|x;\mathcal{E})}, 1-\epsilon, 1+\epsilon \right) A_i \right)
\right)
- \beta \, \mathbb{D}_{\text{KL}} \left( \pi_\theta \,\|\, \pi_{\theta_{\text{ref}}} \right)
\right],
\end{multline}
} 

where $A_i = \left( R^i - \text{mean}\left(\{R^j\}_{j=1}^G\right) \right) \big/ \text{std}\left(\{R^j\}_{j=1}^G\right)$ denotes the normalized advantage of the $i$-th rollout, with $R^i$ as its reward, $\epsilon$ the clipping ratio, and $\beta$ the KL loss coefficient.  A KL divergence penalty is included to constrain the policy $\pi_{\theta_{\text{old}}}$ from straying too far from the reference policy $\pi_{\theta_{\text{ref}}}$.  A key challenge is that execution feedback are externally generated and should not influence gradients.  To ensure correct GRPO optimization, we mask out execution feedback tokens during loss computation, so that updates depend only on model-generated outputs.

\subsection{Rollout with Stepwise SQL Execution Feedback} \label{section_rollout}

To address the limitations of standard reasoning in Text-to-SQL, we propose a stepwise rollout mechanism based on execution-aware reasoning, which integrates SQL execution feedback directly into the generation process. As shown in Appendix~\ref{app:prompt}, we use a structured prompting format with markup tags that delineate model reasoning and database interaction: \textcolor[HTML]{2e57d0}{\texttt{<think>}}...\textcolor[HTML]{2e57d0}{\texttt{</think>}} for reasoning steps, \textcolor[HTML]{ffb713}{\texttt{<intermediate\_sql>}}...\textcolor[HTML]{ffb713}{\texttt{</intermediate\_sql>}} for exploratory SQL queries, and \textcolor[HTML]{16b69e}{\texttt{<result>}}...\textcolor[HTML]{16b69e}{\texttt{</result>}} for execution feedback . Each query is executed by a SQL executor $\mathcal{E}$, and the feedback is appended to the context to guide subsequent reasoning.  The feedback includes either execution errors or query results, where results consist of column names and cell values.  This process repeats until a final SQL query, enclosed in \textcolor[HTML]{bf0040}{\texttt{<final\_sql>}}...\textcolor[HTML]{bf0040}{\texttt{</final\_sql>}}, is produced or a predefined interaction limit $N$ is reached. The full procedure is formalized in Algorithm~\ref{alg:template_rollout}. Stepwise feedback enables progressive refinement of reasoning and improved SQL generation.
\setlength{\algomargin}{1em}       
\SetInd{1.5em}{0.5em}                  
\DontPrintSemicolon
\SetKw{Break}{break}
\SetKw{Continue}{continue}
\begin{algorithm}[h]
\caption{LLM Response Rollout with Stepwise SQL Execution Feedback} 
\label{alg:template_rollout}
\KwIn{Question $x$, database $DB$, policy model $\pi_{\theta}$, SQL executor $\mathcal{E}$, maximum interactions $N$}
\KwOut{Final response $y$, final SQL $s_{\text{final}}$}
Initialize the model output $y \gets \varnothing $ \\
Initialize interactions $n \gets 0 $\\

\While{$n < N$}{
    Initialize the current round model output tokens $y_{\text{temp}} \gets \varnothing $ \\
    \While{True}{
        Generate the next tokens fragment $y_{t} \gets \pi_{\theta}(\cdot|x, y + y_{\text{temp}})$ \\
        $y_{\text{temp}} \gets y_{\text{temp}} + y_{t}$ \\
        \If{$y_{t}$ \emph{in [\textcolor[HTML]{ffb713}{</intermediate\_sql>}, <im\_end>]}}{
            \Break
        }
    }
    $y \gets y + y_{\text{temp}}$ \\
    \If{\emph{\textcolor[HTML]{ffb713}{<intermediate\_sql>}...\textcolor[HTML]{ffb713}{</intermediate\_sql>} in} $y_{\text{\text{temp}}}$}{
        Intermediate SQL query $s_{\text{explore}} \gets $  \\
        {Extract}($y_{\text{temp}}$, pattern="{\textcolor[HTML]{ffb713}{<intermediate\_sql>}...\textcolor[HTML]{ffb713}{</intermediate\_sql>}}") \\
        Execution Feedback $e \gets \mathcal{E}(s_{\text{explore}}, DB)$ \\
        $y \gets y + $ {\textcolor[HTML]{16b69e}{<result>}} $e$ {\textcolor[HTML]{16b69e}{</result>}}
        
    }
    \Else{
        \Break
    }
    Number of interactions $n \gets n+1 $
}
Final SQL $s_{\text{final}} \gets $ {Extract}($y$, pattern="{\textcolor[HTML]{bf0040}{<final\_sql>}...\textcolor[HTML]{bf0040}{</final\_sql>}}") \\
\KwRet Final response $y$, Final SQL $s_{\text{final}}$
\end{algorithm}
\subsection{Composite Execution-Aware Reward Design}

To effectively guide policy optimization, we design a composite reward function that captures multiple aspects of high-quality SQL generation. Specifically, our reward consists of five components—format, execution, exact match, entity match, and exploration—each offering complementary supervision to steer the model toward producing correct, executable, and semantically faithful SQL queries.

\textbf{Format reward}\hspace{0.3cm} The format reward $R_{\text{format}}$ encourages structured reasoning and efficient database interaction by enforcing strict output formatting. It assigns a reward of 1 only when the output $y$ perfectly conforms to the predefined template, and 0 otherwise.  This guides the model to internalize the expected format, improving consistency and reducing formatting errors during training.

\textbf{Exact match reward}\hspace{0.3cm} The exact match reward $R_{\text{em}}$ enforces strict correctness by assigning a reward of 1 only when the output SQL $s_{\text{final}}$ exactly matches the reference $s_{\text{gold}}$, and 0 otherwise. This guides the model to generate $s_{\text{final}}$ that is syntactically identical to $s_{\text{gold}}$.

\textbf{Execution reward}\hspace{0.3cm} The execution reward $R_{\text{exec}}$ promotes both syntactic validity and semantic correctness of $s_{\text{final}}$. It consists of two components: a syntax-level reward $R_{\text{valid}}$, assigned as 1 if $s_{\text{final}}$ executes without error and 0 otherwise; and a result-level reward, also 1 if $\text{Exec}(s_{\text{final}}) = \text{Exec}(s_{\text{gold}})$, and 0 otherwise. Here, $\text{Exec}(s)$ denotes the result of executing SQL $s$ on the database.
\begin{equation}
R_{\text{exec}} =
\begin{cases}
R_{\text{valid}}(s_{\text{final}}) + \mathbf{1}\left[\text{Exec}(s_{\text{final}}) = \text{Exec}(s_{\text{gold}})\right], & \text{if } R_{\text{valid}}(s_{\text{final}}) \neq 0 \\
0, & \text{otherwise}
\end{cases}
.
\end{equation}
\textbf{Entity match reward}\hspace{0.3cm} The entity match reward $R_{\text{entity}}$ encourages accurate use of schema entities in $s_{\text{final}}$.  It compares the sets of tables and columns used in $s_{\text{final}}$ and $s_{\text{gold}}$, and computes the reward based on their overlap. Let $\text{Entities}(s)$ denote the set of table and column in SQL $s$.
\begin{equation}
R_{\text{entity}} = \frac{|\text{Entities}(s_{\text{final}}) \cap \text{Entities}(s_{\text{gold}})|}{|\text{Entities}(s_{\text{gold}})|}
.
\end{equation}
\textbf{Exploration Reward}\hspace{0.3cm}  The exploration reward $R_{\text{expl}}$ encourages effective database interaction during reasoning via two mechanisms: (1) penalizing duplicate intermediate SQL $s_{\text{explore}}$ to discourage redundancy, and (2) rewarding broader exploration—measured by a sigmoid function $\sigma(n)$ over the number of interaction steps $n$—when the final SQL is incorrect, so that limited exploration yields lower reward.  Let \( \mathcal{S}_{\text{explore}} \) denote the list of all \( s_{\text{explore}} \) generated during reasoning.
\begin{equation}
R_{\text{expl}} = 
\begin{cases}
0, & \text{if } |\mathcal{S}_{\text{explore}}| \ne |\text{set}(\mathcal{S}_{\text{explore}})| \\
1, & \text{if } R_{\text{exec}} = R_{\text{exec}}^{\text{max}} \\
\sigma(n), & \text{otherwise}
\end{cases}
.
\end{equation}
\textbf{Composite Reward}\hspace{0.3cm} The final reward $R = \textstyle\sum_{i \in \mathcal{C}} w_i R_i $, where $ \mathcal{C} = \{\text{format}, \text{em}, \text{exec}, \text{entity}, \text{expl} \} $ and $ w_i $ is the weight of component $R_i$. Each term provides distinct supervision to promote syntactic validity, execution correctness, schema alignment, and reasoning quality.

\begin{figure}[t]
    \includegraphics[width=1\columnwidth]{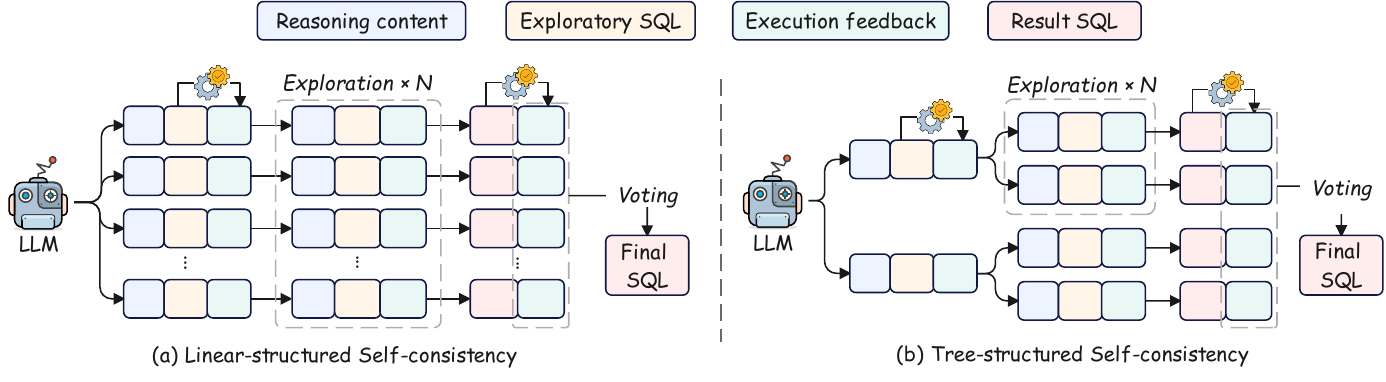} 
    \caption{Comparison between linear-structured and tree-structured self-consistency decoding.}
    \label{fig:tree_decoding}
\end{figure}

\subsection{Tree-Structured Execution-Guided Decoding}

Unlike linear-structured decoding, ReEx-SQL introduces a tree-structured decoding paradigm that explores a broader reasoning space, as illustrated in Figure~\ref{fig:tree_decoding}. Each node in the decoding tree represents the current reasoning context, an exploratory SQL query, and its associated execution feedback. Child nodes correspond to plausible continuations of the reasoning process, enabling a dynamic and adaptive exploration of the reasoning path.  The decoding process adapts iteratively, guided by execution feedback, until a final SQL query is produced or the computational budget is exhausted. After generating multiple SQL candidates, ReEx-SQL employs an execution-based self-consistency mechanism to select the most semantically consistent SQL as the final output. By integrating real-time execution feedback into the generation process, ReEx-SQL achieves more flexible and robust reasoning, significantly enhancing the semantic accuracy of the generated SQL.

\section{Experiments}
\subsection{Experimental Settings} \label{section:expsetting}
\paragraph{Datasets and Metrics}
We conduct experiments on two widely used Text-to-SQL benchmarks: BIRD~\citep{DBLP:conf/nips/bird} and Spider~\citep{DBLP:conf/emnlp/spider}. BIRD is a large-scale cross-domain dataset comprising over 12,000 queries across 95 databases spanning 37 professional domains, designed to better reflect real-world scenarios. The BIRD development set (hereafter BIRD Dev) is evaluated using the official metrics: execution accuracy (EX) and valid efficiency score (VES), where VES measures the runtime efficiency of the predicted SQL compared to the ground truth. Spider, the standard benchmark for cross-domain generalization, includes 200 databases across 138 domains.  We use its development set (hereafter Spider Dev) for evaluation following the official evaluation protocol~\citep{DBLP:conf/emnlp/spider}.  In addition to Spider, we evaluate three of its robustness variants—Spider-Syn~\citep{DBLP:conf/acl/Spider-Syn}, Spider-Realistic~\citep{DBLP:conf/naacl/Spider-Realistic}, and Spider-DK~\citep{DBLP:conf/emnlp/Spider-DK}—following the official evaluation protocol. For Spider and its variants, we use execution accuracy (EX) and test-suite accuracy (TS)~\citep{Test-Suites}. TS assesses whether a SQL query produces correct results across multiple test cases in a test suite of databases, with augmented test suite of database entries ensuring broader coverage. Specifically, Spider-Syn replaces schema-related words with synonyms, Spider-Realistic removes explicitly mentioned column names from the questions, and Spider-DK incorporates domain knowledge to rephrase the questions.

\paragraph{Implementation Details}
We use Qwen2.5-Coder-7B-Instruct model for experiments. During both training and inference, we adopt database prompts from CodeS~\citep{DBLP:journals/pacmmod/codes}, which provide filtered schema components, values, and metadata, and have demonstrated competitive performance on the BIRD benchmark~\citep{DBLP:conf/nips/bird,Schema_Linking_Knapsack}. We use the GRPO~\citep{GRPO} within the OpenRLHF~\citep{OpenRLHF} framework. The training is configured with a batch size of 64 and learning rate of 2e-6. In the rollout phase, we sample 8 outputs per input with a temperature $T$ of 1.0, set the maximum sequence length to 4096, and up to maximum interactions $N{=}10$.  The reward weight vector is defined as $
\mathbf{w} = (w_\text{format}, w_\text{em}, w_\text{exec}, w_\text{entity}, w_\text{expl}) = (2.0, 1.0, 3.0, 1.0, 2.0)$, a configuration referred to as \textbf{MaxTune}. During inference, we apply greedy decoding ($T{=}0.0$), and tree-structured decoding ($T{=}0.7$), allowing up to 3 children per node and generating up to 16 candidates.  We use SQLite~\citep{sqlite} as the SQL executor to obtain execution feedback. The feedback includes column headers and up to 3 rows of cell values.  All experiments are conducted on a system with a 24-core CPU at 2.10GHz, and 8 NVIDIA A800 GPUs.

\begin{table}[t]
\begin{minipage}{0.51\linewidth}
\caption{Performance comparison on BIRD and Spider benchmarks (\%).}
\label{table:exp1_indomain}
\scriptsize 
\setlength{\tabcolsep}{2.2pt}  
\begin{tabular}{lccccc}
\toprule
\multirow{2}{*}{ Methods} & \multicolumn{2}{c}{ BIRD Dev }& \multicolumn{2}{c}{ Spider Dev } & \multicolumn{1}{c}{Spider Test}\\
    \cmidrule(r){2-3} \cmidrule(l){4-5} \cmidrule(l){6-6} 
    & EX  & VES & EX & TS &  EX \\
\midrule
XiYan-SQL~\citep{xiyansql} & \textbf{73.3} & - & - & - & \textbf{89.6} \\
CHASE-SQL + Gemini 1.5~\citep{pourreza2024chase} & 73.0 & \textbf{73.0} & - & - & 87.6 \\
MCTS-SQL + GPT-4~\citep{MCTS-SQL} & 69.4 & 66.2 & 88.7 & - & 86.6 \\
PAS-SQL + GPT-4o~\citep{PAS-SQL} & 64.7 & 65.0 & 87.9 & - & 86.8 \\
MCS-SQL + GPT-4~\citep{mcssql} & 63.4 & 64.8 & \textbf{89.5} & - & \textbf{89.6} \\
MAC-SQL + GPT-4~\citep{DBLP:journals/corr/macsql} & 59.4 & 66.2 & 86.8 & \textbf{82.8} & - \\
DAIL-SQL + GPT-4~\citep{DBLP:journals/pvldb/dail} & 54.8 & 56.1 & 83.5 & 76.2 & - \\
\midrule
SENSE-13B~\citep{DBLP:SENSE} & 55.5 & - & 84.1 & 83.5 & 86.6 \\
DPO + Qwen2.5-Coder-7B~\citep{Dpo-sql} & 64.1 & - & 82.6 & 80.2 & - \\
SFT CodeS-7B~\citep{DBLP:journals/pacmmod/codes} & 57.2 & 58.8 & 85.4 & 80.3 & - \\
DTS-SQL + DeepSeek-7B~\citep{DTS-SQL} & 55.8 & 60.3 & 82.7 & 78.4 & - \\
DB-Explore-7B~\citep{DB-Explore} & 52.1 & 55.8 & 84 & 79.3 & - \\
ROUTE + Qwen2.5-7B~\citep{ROUTE} & 55.9 & 57.4 & 83.6 & 77.5 & 83.7 \\
ROUTE + Qwen2.5-14B~\citep{ROUTE} & 60.8 & 65.2 & 87.3 & 80.9 & \textbf{87.1} \\
SQL-o1 + Llama3-8B~\citep{SQL-o1} & 63.4 & 64.7 & 87.4 & 79.6 & 85.4 \\
\midrule
ReEx-SQL-7B (\textcolor{red}{\textbf{Our}}) & \textbf{64.9} & \textbf{73.1} & \textbf{88.8} & \textbf{83.7} & 86.6 \\
\bottomrule
\end{tabular}
\end{minipage}
\hfil
\begin{minipage}{0.44\linewidth}
\centering
\caption{Performance comparison on Spider variants robustness benchmarks (\%).}
\label{table:exp1_outdomain}
\centering
\scriptsize 
\setlength{\tabcolsep}{2.2pt}  
\begin{tabular}{lccccc}
\toprule
\multirow{2}{*}{Methods} & \multicolumn{2}{c}{Syn} & \multicolumn{2}{c}{Realistic} & \multicolumn{1}{c}{DK}\\
    \cmidrule(r){2-3} \cmidrule(l){4-5} \cmidrule(l){6-6}
    & EX  & TS  & EX & TS& EX \\
    \midrule
FastRAT\textsubscript{ext}+GPT-4~\citep{fastrat} & 74.4 & - & \textbf{80.9} & - & 72.3 \\
TA-SQL + GPT-4~\citep{TASQL} & - & - & 79.5 & - & 72.9 \\
CYCLESQL + GPT-4~\citep{CYCLESQL} & \textbf{76.0} & \textbf{66.3} & 70.6 & 56.9 & \textbf{68.5} \\
SQL-PaLM + PaLM2~\citep{SQLpalm}  & 74.6 & - & 77.6 & - & 66.5 \\
\midrule
DPO + Qwen2.5-Coder-7B~\citep{Dpo-sql} & 76.2 & - & 79.1 & - & 72.9 \\
Qwen2.5-Coder-32B~\citep{OmniSQL} & 70.5 & - & 74.8 & - & 78.3 \\
ROUTE + Llama3-8B~\citep{ROUTE} & 77.4 & 70.2 & 80.9 & 72.6 & 74.6 \\
SFT CodeS-7B~\citep{DBLP:journals/pacmmod/codes} & 76.9 & 70.0 & 82.9 & 77.2 & 72.0 \\
SFT CodeS-15B~\citep{DBLP:journals/pacmmod/codes} & 77.0 & 69.4 & 83.1 & 75.6 & 70.7 \\
SENSE-7B~\citep{DBLP:SENSE} & 72.6 & 64.9 & 82.7 & 75.6 & 77.9 \\
SENSE-13B~\citep{DBLP:SENSE} & 77.6 & 70.2 & 84.1 & 76.6 & \textbf{80.2} \\
OmniSQL-7B~\citep{OmniSQL} & 69.6 & - & 78.0 & - & 77.8 \\
OmniSQL-14B~\citep{OmniSQL} & 72.0 & - & 78.5 & - & 74.8 \\
OmniSQL-32B~\citep{OmniSQL} & 72.1 & - & 77.2 & - & 77.6 \\
SQL-o1 + Llama3-8B~\citep{SQL-o1} & 77.6 & 69.2 & 82.7 & 72.8 & 78.7 \\
\midrule
ReEx-SQL-7B (\textcolor{red}{\textbf{Our}}) & \textbf{78.6} & \textbf{72.1} & \textbf{85.2} & \textbf{79.1} & 79.8 \\
\bottomrule
\end{tabular}
\end{minipage}
\end{table}

\subsection{Overall Performance} \label{exp:Overall}

\paragraph{Results on BIRD and Spider}



Table~\ref{table:exp1_indomain} presents the performance of ReEx-SQL on the BIRD and Spider benchmarks. The results demonstrate that ReEx-SQL achieves competitive performance under the same model size. On the BIRD Dev, it reaches an execution accuracy of 64.9\%, significantly narrowing the gap with proprietary solutions. Moreover, it achieves a VES score of 73.1\%, indicating better execution effectiveness of the generated SQL. On the Spider Dev, ReEx-SQL attains a TS score of 83.7\%, outperforming several proprietary and large-scale models, which shows its strong capability in generating semantically accurate SQL.  Although ReEx-SQL lags behind recent SOTA models~\citep{xiyansql,pourreza2024chase}, these approaches often come with high costs—XiYan-SQL~\citep{xiyansql} uses multiple models, and CHASE-SQL~\citep{pourreza2024chase} relies on the proprietary GPT-4. In contrast, ReEx-SQL runs on a single 7B open-source model, offering better efficiency and ease of deployment.


\paragraph{Robustness Evaluation}

Table~\ref{table:exp1_outdomain} shows the robustness evaluation across three Spider variants, highlighting the following:  (1) ReEx-SQL-7B achieves the highest EX scores, with 78.6\% on Spider-Syn and 85.2\% on Spider-Realistic, outperforming both open-source and proprietary models.  (2) It also leads in TS scores, with 72.1\% on Spider-Syn and 79.1\% on Spider-Realistic, indicating better structural alignment and fewer semantic errors.  (3) On Spider-DK, ReEx-SQL achieves a strong score of 79.8\%, using only a 7B model without any additional augmented data. In contrast, SENSE-13B, which benefits from a larger model and extra data augmentation, achieves only a slightly higher score of 80.2\%. This highlights ReEx-SQL’s superior efficiency and generalization under limited-resource settings.  Overall, ReEx-SQL-7B demonstrates strong robustness and generalization across diverse perturbations and knowledge gaps, while remaining lightweight and cost-effective.

\begin{table}[t]
\caption{Comparison of reasoning paradigms on multiple Text-to-SQL benchmarks.}
\label{table:exp2_baseline_and_exec_compare}
\centering
\scriptsize 
\setlength{\tabcolsep}{1.8pt}  
\begin{tabular}{ccccccccccc}
\toprule
\multirow{2}{*}{ \makecell{Reasoning \\Paradigm}} & \multirow{2}{*}{\makecell{Training\\ Type}} & \multicolumn{1}{c}{BIRD Dev }& \multicolumn{2}{c}{ Spider Dev } & \multicolumn{1}{c}{Spider Test} & \multicolumn{2}{c}{Spider-Syn} & \multicolumn{2}{c}{Spider-Realistic} & \multicolumn{1}{c}{Spider-DK}\\
    \cmidrule(r){3-3} \cmidrule(r){4-5} \cmidrule(l){6-6} \cmidrule(l){7-8}  \cmidrule(l){9-10}   \cmidrule(l){11-11} 
    & & EX (\%) & EX (\%) & TS (\%) &  EX (\%)  & EX (\%) & TS (\%)  & EX (\%) & TS (\%)  & EX (\%) \\
\midrule
Direct Output & No Post-Training & 48.7 & 81.4 & 75.3 & 79.3 & 68.3 & 54.6 & 77.0 & 68.4 & 71.6 \\
Standard Reasoning & No Post-Training & 49.5 & 81.1 & 74.8 & 78.3 & 69.5 & 56.9 & 74.6 & 63.1 & 71.4 \\
Execution-Aware Reasoning & No Post-Training & 48.6 & 80.0 & 75.9 & 37.0 & 57.4 & 43.1 & 65.7 & 53.8 & 57.6 \\
Direct Output & SFT & 56.3 & 85.4 & 81.4 & 84.5 & 75.7 & 68.3 & 81.9 & 74.2 & 74.0 \\
\midrule
Standard Reasoning & GRPO & 60.8 & 85.7 & 83.5 & 86.2 & 78.3 & 68.9 & 82.3 & 75.6 & 74.6 \\
\textbf{Execution-Aware Reasoning} & GRPO
& \textbf{63.4\textsubscript{+2.6}} 
& \textbf{88.4\textsubscript{+2.7}} 
& \textbf{83.5\textsubscript{+0.0}} 
& \textbf{86.4\textsubscript{+0.2}} 
& \textbf{78.5\textsubscript{+0.2}} 
& \textbf{72.1\textsubscript{+3.2}} 
& \textbf{85.0\textsubscript{+2.7}} 
& \textbf{78.7\textsubscript{+3.1}} 
& \textbf{79.3\textsubscript{+4.7}} \\
\bottomrule
\end{tabular}
\begin{tablenotes}
\scriptsize
\item * All results are obtained with greedy decoding. ``Direct Output'' refers to directly predicting SQL without intermediate reasoning.
\end{tablenotes}
\end{table}

\begin{figure}[t]
    \captionsetup{aboveskip=2pt}
    \includegraphics[width=1\columnwidth]{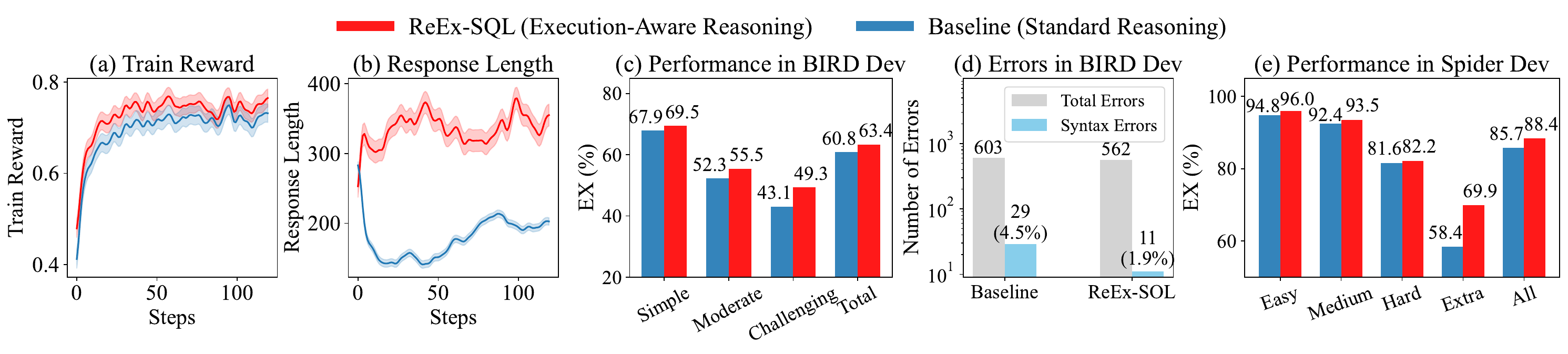} 
    \caption{
    Multi-dimensional comparison between ReEx-SQL and the baseline: (a) Training reward over steps; (b) Response length throughout training; (c) BIRD performance across difficulty levels; (d) Error type distribution on the BIRD Dev; (e) Spider performance across difficulty levels.
    }
    \label{fig:exp2_baseline_and_exec_compare}
\end{figure}

\subsection{Ablation Study on Reasoning Paradigms} \label{exp:compare}

To highlight the strength of the execution-aware reasoning paradigm in ReEx-SQL, Table~\ref{table:exp2_baseline_and_exec_compare} compares the performance of different reasoning paradigms. For simplicity, we refer to the GRPO-trained method under the standard reasoning paradigm as the baseline. To ensure fairness, ReEx-SQL and the baseline use identical training configurations, with two key differences: they adopt different prompting templates, and exploration reward $R_{\text{expl}}$ is applied only in ReEx-SQL. The prompting templates used for each paradigm are provided in Appendix~\ref{app:prompt}.  Highlighting the following: (1) After GRPO training, ReEx-SQL outperforms the baseline across all benchmarks. It achieves a 2.6\% and 2.7\% gain in EX score on BIRD Dev and Spider Dev, respectively, demonstrating strong in-domain performance. For robustness, it improves TS scores on Spider-Syn and Spider-Realistic by 3.2\% and 3.1\%, and surpasses the baseline on Spider-DK by 4.7\% in EX score. These improvements highlight that the advantage lies in the execution-aware reasoning paradigm itself.  (2)  ReEx-SQL improves significantly after GRPO training—from 48.6\% to 63.4\%, a 14.8\% gain compared to its untrained version—while the baseline improves by 11.3\% over its own untrained counterpart. This confirms the effectiveness of GRPO and further suggests that the execution-aware reasoning paradigm unlocks greater potential under GRPO, offering higher ceilings and broader room for improvement.

\paragraph{ReEx-SQL vs. Baseline} 

To further highlight the advantages of ReEx-SQL, we conduct a more comprehensive comparison with the baseline, as shown in Figure~\ref{fig:exp2_baseline_and_exec_compare}. The key observations are as follows: (1) Training rewards (Figure~\ref{fig:exp2_baseline_and_exec_compare}a): Both methods converge quickly and remain stable during training. However, ReEx-SQL consistently achieves higher average rewards than the baseline, indicating a higher performance ceiling under GRPO training. This aligns with the improvements observed in Table~\ref{table:exp2_baseline_and_exec_compare}.  (2) Response length (Figure~\ref{fig:exp2_baseline_and_exec_compare}b): Both methods show longer reasoning over time, echoing DeepSeek's finding that models learn to think longer~\citep{Deepseek-r1}. ReEx-SQL quickly expands responses by learning to interact, while the baseline first trims noise before reasoning more deeply. (3) Performance by difficulty level (Figures~\ref{fig:exp2_baseline_and_exec_compare}c and e): We divide the BIRD Dev and Spider Dev based on difficulty. ReEx-SQL outperforms the baseline across all difficulty levels, with particularly notable gains on challenging and extra hard examples: 6.2\% and 11.5\% improvements on BIRD and Spider Dev, respectively. This demonstrates ReEx-SQL’s superior ability to handle complex queries, a strength stemming from the execution-aware reasoning paradigm itself.  (4)  Error analysis (Figure~\ref{fig:exp2_baseline_and_exec_compare}d): On BIRD Dev, ReEx-SQL reduces the total number of errors from 603 to 562 compared to the baseline. Notably, the proportion of syntax errors drops from 4.5\% to 1.9\%, yielding a 2.6\% absolute improvement.  These results indicate that ReEx-SQL effectively mitigates both syntactic and semantic errors, addressing the shortcomings of the baseline illustrated in Figure~\ref{fig:motivation}.  Representative output comparisons are provided in Appendix~\ref{app:output_compare}.

\subsection{Ablation Study on Reward Design} \label{exp:reward_design}

\begin{table}[t]
\begin{minipage}{0.4\linewidth}
\centering

\caption{Ablation of reward components for ReEx-SQL on BIRD Dev.}
\label{table:exp3_reward_wo}
\scriptsize 
\begin{tabular}{lcc}
\toprule
\multirow{2}{*}{Reward Components} & \multicolumn{2}{c}{ BIRD Dev }\\
        \cmidrule(r){2-3}
        & EX (\%) & VES (\%)  \\
\midrule
\makecell{ReEx-SQL \\(Base Configuration)} & \textbf{60.9} & \textbf{70.8} \\
\hdashline 
\hspace{0.3cm} w/o $R_{\text{em}}$ & 58.0 {\tiny (\ensuremath{\downarrow} 2.9)} & 65.4 {\tiny (\ensuremath{\downarrow} 5.4)} \\
\hspace{0.3cm} w/o $R_{\text{exec}}$ & 54.4 {\tiny (\ensuremath{\downarrow} 6.5)} & 62.7 {\tiny (\ensuremath{\downarrow} 8.1)}\\
\hspace{0.3cm} w/o $R_{\text{entity}}$ & 57.8 {\tiny (\ensuremath{\downarrow} 3.1)} & 66.1 {\tiny (\ensuremath{\downarrow} 4.7)}\\
\hspace{0.3cm} w/o $R_{\text{expl}}$ & 59.2 {\tiny (\ensuremath{\downarrow} 1.7)} & 69.3 {\tiny (\ensuremath{\downarrow} 1.5)}\\
\bottomrule
\end{tabular}
\end{minipage}
\hfil
\begin{minipage}{0.54\linewidth}
\centering
\caption{Ablation of reward weights for ReEx-SQL on BIRD Dev.  Configuration names reflect the characteristic reward weight assignments.}
\label{table:exp3_reward_weight}
\centering
\scriptsize 
\setlength{\tabcolsep}{2.5pt}  
\begin{tabular}{cccccccc}
\toprule
\multirow{2}{*}{ \makecell{Configuration\\ Type}} &  \multicolumn{5}{c}{Reward Weights} & \multicolumn{2}{c}{ BIRD Dev }\\
\cmidrule(r){2-6} \cmidrule(r){7-8}
 & $w_\text{format}$ & $w_\text{em}$ & $w_\text{exec}$ & $w_\text{entity}$ & $w_\text{expl}$ & EX (\%) & VES (\%) \\
\midrule
Base & 1.5 & 0.8 & 1.5 & 0.8 & 0.5 & 60.9 & 70.8 \\
HighExploration & 1.5 & 0.8 & 1.5 & 0.8 & 2.0 & 62.0 & 69.4 \\
HighFormat-EX & 2.0 & 0.8 & 3.0 & 0.8 & 0.5 & 62.3 & 70.1 \\
Uniform & 1.0 & 1.0 & 1.0 & 1.0 & 1.0 & 62.3 & \textbf{73.8} \\
MaxTune (\textbf{Best}) & 2.0 & 1.0 & 3.0 & 1.0 & 2.0 & \textbf{63.4} & 72.3 \\
\bottomrule
\end{tabular}
\end{minipage}
\end{table}

\paragraph{Ablation of Reward Components}
Table~\ref{table:exp3_reward_wo} presents the impact of each reward component, with all results based on configuration ``Base'', as detailed in Table~\ref{table:exp3_reward_weight}.  The results show that removing any reward leads to a performance drop in ReEx-SQL, highlighting the necessity of each component.  Execution reward $R_{\text{exec}}$ has the most significant impact, highlighting the unique advantage of leveraging execution feedback in the Text-to-SQL.  Although the exploration reward presents a relatively smaller direct effect,  as discussed in Appendix~\ref{app:reward_design}, it substantially enriches exploration diversity, ensuring that each interaction yields informative gains.

\paragraph{Ablation of Reward Weights}
Table~\ref{table:exp3_reward_weight} presents the impact of different reward weight configurations on ReEx-SQL, with the following key observations:  (1) Increasing the exploration reward  weight $w_{\text{expl}}$ (Base vs. HighExploration) leads to a 1.1\% improvement in EX score on BIRD Dev (60.9\% vs. 62.0\%), suggesting that encouraging meaningful interactions with the database can facilitate the generation of higher-quality SQL.  (2) Increasing the weights for format $w_{\text{format}}$ and execution $w_{\text{exec}}$ (Base vs. HighFormat-EX) improves the EX score by 1.4\%, indicating that enforcing consistency in output format and execution results helps the model generate better-structured and semantically aligned SQL.  (3)  MaxTune, which simultaneously increases $w_{\text{format}}$, $w_{\text{exec}}$, and $w_{\text{expl}}$, achieves the best performance, highlighting the synergistic effect of combining multiple reward signals.

\begin{table}[t]
\caption{Comparison of decoding strategies for ReEx-SQL on BIRD and Spider benchmarks. }
\label{table:exp4_comparsion_decoding}
\centering
\scriptsize 
\setlength{\tabcolsep}{5pt}  
\begin{tabular}{lccccccc}
\toprule
\multirow{2}{*}{Decoding Strategy} &\multirow{2}{*}{\makecell{Number of\\Candidates} } & \multicolumn{3}{c}{BIRD Dev}  & \multicolumn{3}{c}{ Spider Dev} \\
\cmidrule(r){3-5} \cmidrule(r){6-8}
 & & EX(\%) & VES(\%) & Time (s/sample) \ensuremath{\downarrow} & EX(\%) & TS(\%) & Time (s/sample) \ensuremath{\downarrow} \\
\midrule
Greedy decoding& 1 & 63.4 & 72.7 & 0.6 & 88.4 & 83.5 & 0.2 \\
\midrule
Linear-structured self-consistency& 8 & 64.1 & \textbf{72.7} & 2.8 & 88.3 & 83.6 & 1.4 \\
Tree-structured self-consistency & 8  & \textbf{64.2} & 72.2 & \textbf{2.0 {\scriptsize (\ensuremath{\downarrow} 28.6\%)}} & \textbf{88.4} & \textbf{83.6} & \textbf{0.9 {\scriptsize (\ensuremath{\downarrow} 35.7\%)}} \\
\midrule
Linear-structured self-consistency & 16 & 64.5 & 72.9 & 5.4 & 88.5 & 83.5 & 2.8 \\
Tree-structured self-consistency & 16 & \textbf{64.9} & \textbf{73.1} & \textbf{2.6 {\scriptsize (\ensuremath{\downarrow} 51.9\%)}} & \textbf{88.8} & \textbf{83.7} & \textbf{1.1 {\scriptsize (\ensuremath{\downarrow} 60.7\%)}} \\
\bottomrule
\end{tabular}
\end{table}

\subsection{Decoding Strategy Comparison}

Table~\ref{table:exp4_comparsion_decoding} presents the impact of different decoding strategies on ReEx-SQL. With the same number of candidates, tree-structured decoding consistently outperforms its linear-structured counterpart. When the candidate size is 16, it improves the EX score by 0.4\% on BIRD Dev and 0.3\% on Spider Dev, suggesting its superior ability to explore the underlying reasoning space and generate higher-quality SQL.  In addition to performance gains, tree-structured decoding significantly improves inference efficiency. It reduces average decoding time per sample by 60.7\% on Spider Dev (1.1s vs. 2.8s) and by 51.9\% on BIRD Dev (2.6s vs. 5.4s).  These results demonstrate that Tree-structured decoding is not only more effective but also more efficient, making it particularly suitable for training scenarios with frequent interactions or real-time applications.  For a more comprehensive evaluation, Appendix~\ref{decoding_strategy_spider_variants} presents decoding results on Spider variants.


\subsection{Impact of Execution Feedback and Database}

\begin{table}[t]
\caption{Impact of execution feedback and database complexity on ReEx-SQL.}
\label{table:exp5_execution_database_impact}
\centering
\scriptsize 
\setlength{\tabcolsep}{3.2pt}  
\begin{tabular}{cccccc|ccccc|ccccc}
\toprule
\multirow{2}{*}{Database} & \multirow{2}{*}{Nums} & \multicolumn{4}{c}{BIRD Dev (EX\%) }& \multicolumn{5}{c}{ Spider Test (EX\%)} & \multicolumn{5}{c}{Spider-DK (EX\%)} \\
\cmidrule(r){3-6} \cmidrule(r){7-11} \cmidrule(r){12-16} 
 &  & Simple & Moderate & Challenging & Total & Easy & Medium & Hard & Extra & All & Easy & Medium & Hard & Extra & All \\
\midrule
Origin & 3 & \textbf{69.5} & 55.5 & 49.3 & 63.4 & 93.8 & 89.0 & 82.1 & 76.2 & 86.4 & 88.2 & 85.4 & 62.2 & 67.6 & 79.3 \\
Origin & 6 & 69.4 & 55.7 & 49.3 & 63.4 & \textbf{94.0} & 89.0 & 82.3 & \textbf{76.5} & 86.6 & 88.2 & 85.8 & 64.9 & 66.7 & 79.6 \\
Origin & 8 & 69.4 & \textbf{55.9} & \textbf{50.0} & \textbf{63.5} & 93.8 & 89.0 & 82.3 & 76.2 & 86.5 & 88.2 & 85.8 & 64.9 & \textbf{67.8} & 79.8 \\
\midrule
TS-DB & 3 & - & - & - & - & \textbf{94.0} & 89.0 & \textbf{82.7} & \textbf{76.5} & \textbf{86.7 }& 88.2 & \textbf{86.2} & 64.9 & 67.6 & \textbf{80.0} \\

\bottomrule
\end{tabular}
\begin{tablenotes}
\scriptsize
\item * All results are with greedy decoding. TS-DB refers to the augmented databases for TS score evaluation~\citep{Test-Suites}, from Spider and its variants.
\end{tablenotes}
\end{table}

To investigate how the richness of execution feedback affects performance, we compare different result row limits and database settings, as shown in Table~\ref{table:exp5_execution_database_impact}.  ``Origin'' refers to the default evaluation database, while TS-DB contains more diverse cell values. ``Nums'' indicates the number of rows returned in each execution result.  Results show that increasing the number of feedback rows improves performance—for example, raising Nums from 3 to 8 boosts the EX score on BIRD Challenging by 0.7\% (49.3\% vs. 50.0\%).  Additionally, using TS-DB leads to further gains, such as a 0.6\% improvement on Spider Test Hard (82.1\% vs. 82.7\%) and a 0.7\% gain on Spider-DK (79.3\% vs. 80.0\%). These findings indicate that richer feedback provides more informative guidance for reasoning, resulting in higher-quality SQL generation, particularly for complex queries.

\section{Conclusion} \label{conclusion}


We present ReEx-SQL, a novel framework that introduces an execution-aware reasoning paradigm by integrating real-time execution feedback into the decoding process. Leveraging a structured prompting format and iterative interaction with the database, ReEx-SQL enables the model to progressively refine its reasoning paths. To enhance learning, we extend GRPO with execution-aware rollouts and design a composite reward function that supervises SQL quality across format, execution, semantic correctness, and exploration. Experiments demonstrate strong in-domain and out-of-domain performance, consistently surpassing similarly sized models and some proprietary methods. Furthermore, we introduce a tree-structured decoding strategy that reduces inference time while maintaining or improving accuracy under the same candidate budget, highlighting the efficiency of execution-guided reasoning. Overall, ReEx-SQL shows that execution feedback can effectively guide reasoning, improving both performance and generalization of small open-source LLMs, and narrowing the gap with proprietary solutions. 


While ReEx-SQL effectively improves reasoning via execution feedback, its reliance on database access may limit applicability in certain scenarios, such as offline or privacy-restricted environments. Future work could explore lightweight execution simulators or fallback mechanisms to address this. Additionally, more advanced decoding strategies may further expand the reasoning space and improve efficiency under constrained settings.



{
\small
\bibliographystyle{unsrtnat}
\bibliography{neurips_2025}
}

\newpage
\appendix

\section{Appendices}


\subsection{Prompt Templates} \label{app:prompt}

This section presents the prompts corresponding to the three decoding strategies evaluated in Table~\ref{table:exp2_baseline_and_exec_compare}. The execution-aware reasoning prompt, shown in Figure~\ref{prompt:execprompt}, explicitly incorporates execution feedback to guide the generation process; this prompt is used by ReEx-SQL. The standard reasoning prompt, shown in Figure~\ref{prompt:woexecprompt},performs multi-step reasoning without relying on execution signals, and serves as the decoding strategy for the baseline. In contrast, the direct output prompt in Figure~\ref{prompt:sftprompt} generates SQL queries directly, without any intermediate reasoning steps.

\begin{figure}[H]
\begin{tcolorbox}[title={Execution-Aware Reasoning Prompt},colback=yellow!1.5] 
You are an experienced database expert. Now you need to generate a SQL query given the database information, a question, and some additional information.
Your goal is to generate a single **SQLite** query that can correctly answer the user's question based on the given schema and matched values.\\
\rule{\linewidth}{0.4pt} 

\textbf{Important Guidelines:}
\begin{enumerate}[leftmargin=1em]
  \item The database structure is defined by the following table schemas:  \\
    **table\_name.column\_name ( data\_type | comment : description | values : values )**  \\
    The ``values'' are only examples to illustrate the data type and format; They are not directly related to the question.
  \item You should:
      \begin{enumerate}[label=\arabic{enumi}.\arabic*.]
      \item **Analyze** the question intent and map it to relevant tables and columns.
      \item If the question is complex or ambiguous, you may write an **intermediate SQL** for verification.
      \item **Do not generate an infinite number of intermediate SQLs**. Instead, focus on identifying key areas where the SQL might need improvement and stop when you have sufficient insight.
      \item When reflecting on the intermediate SQL:
      \begin{itemize}
      \item First, check whether it **correctly expresses all conditions, columns, and logic** described in the original question.
      \item Also, verify whether it follows the Database admin instructions, such as avoiding prohibited patterns or optimizing structure.
      \item Then, evaluate whether the **execution result** confirms or contradicts your expectations.
      \item If there is any mismatch, misunderstanding, violation of admin rules, or missing condition, **revise your reasoning before generating the final SQL**.
      \item In addition, you can generate other possible intermediate SQLs for comparison.
      \end{itemize}
    \end{enumerate}
\end{enumerate}
\rule{\linewidth}{0.4pt} 

\textbf{Response Format:} \\
Respond strictly in the following format:\\
\textcolor[HTML]{2e57d0}{{<think>}} Reasoning process here. \textcolor[HTML]{2e57d0}{{</think>}} \\
\textcolor[HTML]{ffb713}{{<intermediate\_sql>}} The intermediate SQL query for verification is: ```sql \\
your intermediate SQL here.'''\textcolor[HTML]{ffb713}{{</intermediate\_sql>}} \\
\textcolor[HTML]{16b69e}{{<result>}} Execution result here. \textcolor[HTML]{16b69e}{{</result>}} \\
\textcolor[HTML]{2e57d0}{{<think>}} Further reasoning. \textcolor[HTML]{2e57d0}{{</think>}}  \\
\textcolor[HTML]{bf0040}{{<final\_sql>}} The final SQL query is: ```sql your final SQL here. '''\textcolor[HTML]{bf0040}{{</final\_sql>}} \\
\rule{\linewidth}{0.4pt} 
\textbf{[Table creation statements]} \{database\_schema\}\\
\textbf{[Matched Values]} \{matched\_contents\}\\
\textbf{[Question]} \{question\}  Hint: \{evidence\} 
\end{tcolorbox}
\caption{Execution-aware reasoning prompt for ReEx-SQL.}
\label{prompt:execprompt}
\end{figure}

\begin{figure}[H]
\begin{tcolorbox}[title={Standard Reasoning Prompt},colback=yellow!1.5] 
You are an experienced database expert. Now you need to generate a SQL query given the database information, a question, and some additional information.
Your goal is to generate a single **SQLite** query that can correctly answer the user's question based on the given schema and matched values.\\
\rule{\linewidth}{0.4pt} 

\textbf{Important Guidelines:}
\begin{enumerate}[leftmargin=1em]
  \item The database structure is defined by the following table schemas:  \\
    **table\_name.column\_name ( data\_type | comment : description | values : values )**  \\
    The ``values'' are only examples to illustrate the data type and format; They are not directly related to the question.
  \item **Analyze** the question intent and map it to relevant tables and columns.
  \item First, check whether it **correctly expresses all conditions, columns, and logic** described in the original question.
  \item Also, verify whether it follows the Database admin instructions, such as avoiding prohibited patterns or optimizing structure.
\end{enumerate}
\rule{\linewidth}{0.4pt} 

\textbf{Response Format:} \\
Respond strictly in the following format:\\
\textcolor[HTML]{2e57d0}{{<think>}} Reasoning process here. \textcolor[HTML]{2e57d0}{{</think>}} \\
\textcolor[HTML]{bf0040}{{<final\_sql>}} The final SQL query is: ```sql your final SQL here. '''\textcolor[HTML]{bf0040}{{</final\_sql>}} \\
\rule{\linewidth}{0.4pt} 
\textbf{[Table creation statements]} \{database\_schema\}\\
\textbf{[Matched Values]} \{matched\_contents\}\\
\textbf{[Question]} \{question\}  Hint: \{evidence\} 
\end{tcolorbox}
\caption{Standard reasoning prompt for baseline.}
\label{prompt:woexecprompt}
\end{figure}

\begin{figure}[H]
\begin{tcolorbox}[title={Direct Output Prompt}, colback=yellow!1.5] 
You are an experienced database expert. Now you need to generate a SQL query given the database information, a question, and some additional information.
Your goal is to generate a single **SQLite** query that can correctly answer the user's question based on the given schema and matched values.\\
\rule{\linewidth}{0.4pt} 
\textbf{Important Guidelines:}
\begin{enumerate}[leftmargin=1em]
  \item The database structure is defined by the following table schemas:  \\
    **table\_name.column\_name ( data\_type | comment : description | values : values )**  \\
    The ``values'' are only examples to illustrate the data type and format; They are not directly related to the question.
  \item Your task is to generate the correct SQL query **only**. Do not include any explanation, reasoning, or additional text.
\end{enumerate}
\rule{\linewidth}{0.4pt} 

\textbf{Response Format:} \\
Respond strictly in the following format:\\
\textcolor[HTML]{bf0040}{{<final\_sql>}} The final SQL query is: ```sql your final SQL here. '''\textcolor[HTML]{bf0040}{{</final\_sql>}} \\
\rule{\linewidth}{0.4pt} 
\textbf{[Table creation statements]} \{database\_schema\}\\
\textbf{[Matched Values]} \{matched\_contents\}\\
\textbf{[Question]} \{question\}  Hint: \{evidence\} 
\end{tcolorbox}
\caption{Direct output prompt.}
\label{prompt:sftprompt}
\end{figure}

\subsection{More Details about Experiments}

\subsubsection{Decoding Strategy Comparison on Spider Variants} \label{decoding_strategy_spider_variants}

Table~\ref{table:exp4_comparsion_decoding_other_dataset} extends the comparison in Table~\ref{table:exp4_comparsion_decoding} by evaluating decoding strategies on various Spider robustness benchmarks. Consistent with previous findings, Tree-structured self-consistency continues to outperform or match other strategies in both effectiveness and robustness. It achieves the highest execution accuracy on Spider-Realistic (85.2\%) and Spider-DK (79.8\%), which are more challenging due to their paraphrased questions and unseen schema structures.  Even when using only 8 candidates, the tree-structured approach performs competitively; and with 16 candidates, it yields further improvements across nearly all metrics. These results reinforce that tree-structured decoding generalizes well under distribution shifts and diverse linguistic perturbations, making it a strong choice for real-world deployment scenarios where robustness is critical.

\begin{table}[t]
\caption{Comparison of decoding strategies on spider variants robustness benchmarks.}
\label{table:exp4_comparsion_decoding_other_dataset}
\centering
\footnotesize 
\setlength{\tabcolsep}{2.5pt}  
\begin{tabular}{lccccccc}
\toprule
\multirow{2}{*}{Decoding Strategy} & \multirow{2}{*}{\makecell{Number of\\Candidates} }  & \multicolumn{1}{c}{Spider Test} & \multicolumn{2}{c}{Spider-Syn} & \multicolumn{2}{c}{Spider-Realistic} & \multicolumn{1}{c}{Spider-DK}\\
 \cmidrule(l){3-3} \cmidrule(l){4-5}  \cmidrule(l){6-7}   \cmidrule(l){8-8} 
   & & EX (\%)  & EX (\%) & TS (\%)  & EX (\%) & TS (\%)  & EX (\%) \\
\midrule
Greedy decoding & 1  & 86.4 & 78.5 & 72.1 & 85.0 & 78.7 & 79.3 \\
\midrule
Linear-structured self-consistency& 8 & \textbf{86.6} & 78.3 & 71.6 & 84.6 & 78.5 & 78.9 \\
Tree-structured self-consistency& 8 & 86.4 & 78.3 & \textbf{71.7} & \textbf{85.8} & \textbf{79.5} & \textbf{79.1} \\
\midrule
Linear-structured self-consistency & 16 & 86.5 & 78.5 & 71.7 & 85.0 & 78.7 & 79.4 \\
Tree-structured self-consistency & 16 & \textbf{86.6} &\textbf{ 78.6} & \textbf{72.1} & \textbf{85.2} & \textbf{79.1} & \textbf{79.8} \\
\bottomrule
\end{tabular}
\end{table}

\begin{figure}[t]
    \includegraphics[width=1\columnwidth]{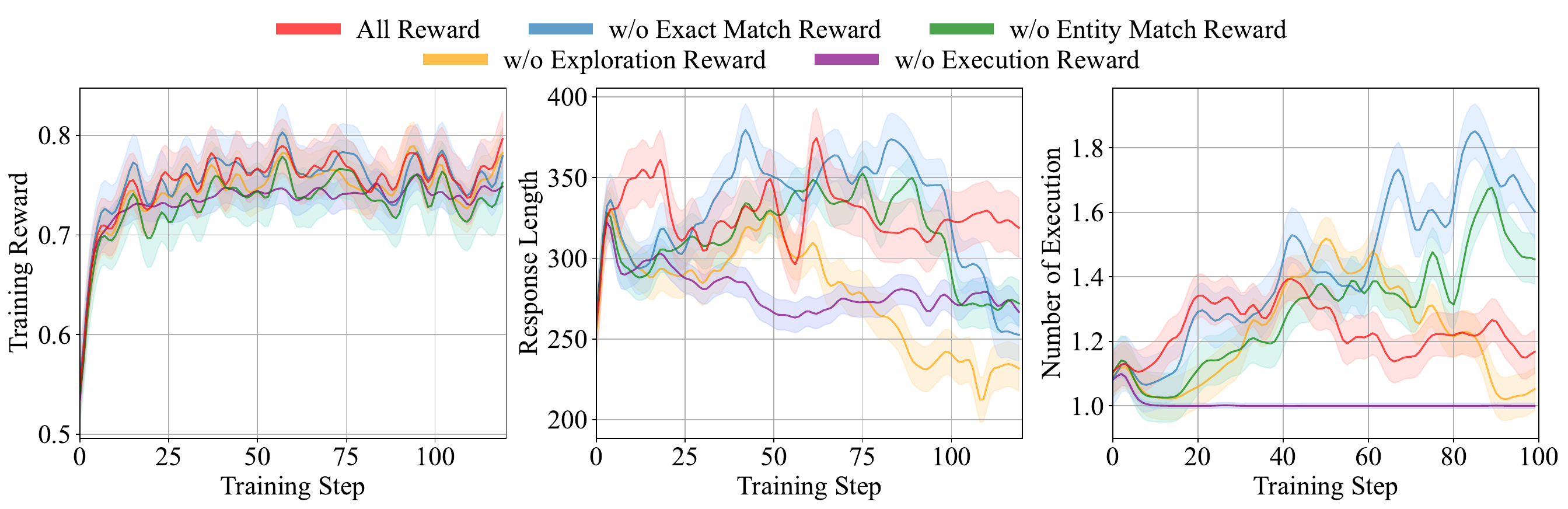} 
    \caption{Impact of reward components on training reward (left), response length (middle), and interaction steps (right).}
    \label{fig:wo_reward}
    
\end{figure}

\begin{figure}[t]
    \includegraphics[width=1\columnwidth]{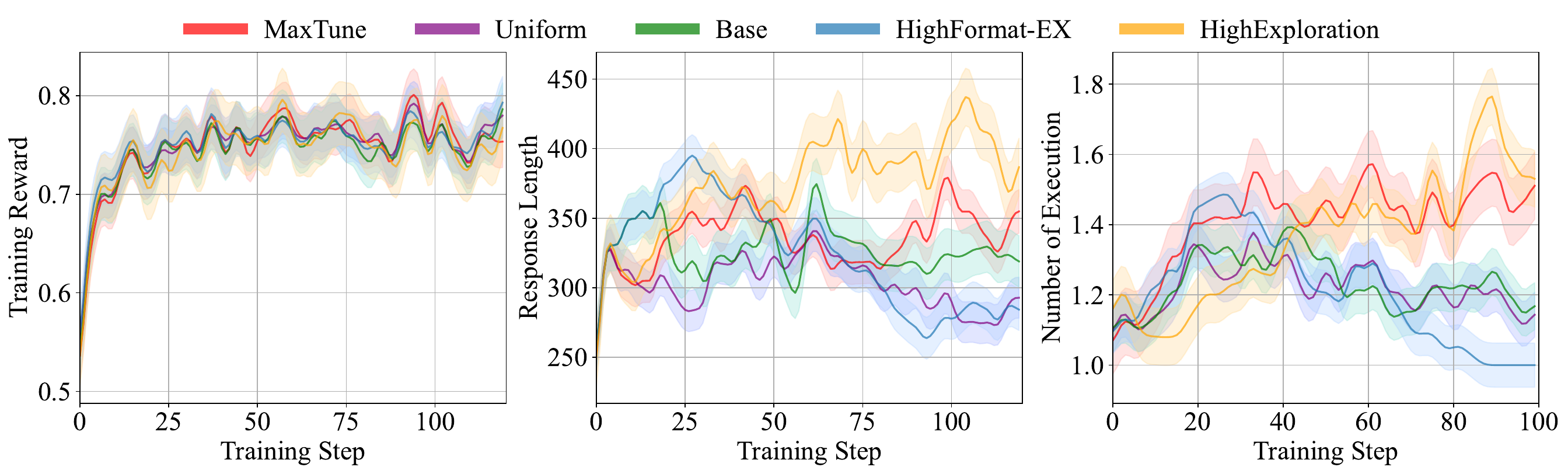} 
    \caption{Impact of different reward weight settings on training reward (left), response length (middle), and interaction steps (right).}
    \label{fig:reward_weight}
\end{figure}

\subsubsection{Impact of Reward Design on Training Dynamics} \label{app:reward_design}

This section complements Section~\ref{exp:reward_design} by analyzing the impact of reward design on training dynamics, focusing on two aspects: reward components and reward weights.

\paragraph{Impact of Reward Components on Training Dynamics}

Figure~\ref{fig:wo_reward} shows how the training reward, response length, and interaction count evolve under different reward ablation settings. We highlight the following observations:  (1) Training converges rapidly within the first 50 steps and remains stable, indicating that GRPO effectively activates model potential and aligns behavior with the reward signal. (2) Without the execution reward, the average reward is significantly lower, consistent with Table~\ref{table:exp3_reward_wo}, highlighting its essential role in the Text-to-SQL task.  (3) Without the exploration reward, response length and interaction count decline in later stages, suggesting the model tends to favor short, low-interaction reasoning paths without explicit exploration incentives.  (4) With all reward components present, the model exhibits stable behavior in both response length and interaction count, indicating a synergistic effect that supports more robust training dynamics.

\paragraph{Impact of Reward Weights on Training Dynamics}

Figure~\ref{fig:reward_weight} illustrates the impact of different reward weight configurations. Key findings include:  (1) Average reward remains similar across configurations, suggesting that the model is robust to moderate changes in reward weights. (2) Increasing the exploration reward weight (HighExploration vs. Base) leads to longer responses and more interactions, showing that stronger incentives promote deeper reasoning.  (3) HighFormat-EX increases the format and execution rewards at the expense of exploration, resulting in shorter responses and fewer interactions, indicating that overemphasis on correctness may suppress exploration.  (4) MaxTune yields intermediate response length and interaction count, reflecting a balanced trade-off between execution accuracy and exploration.

\subsubsection{ReEx-SQL vs. Baseline: Output Analysis} \label{app:output_compare}

\begin{figure}[h]
\centering
    \includegraphics[width=1\columnwidth]{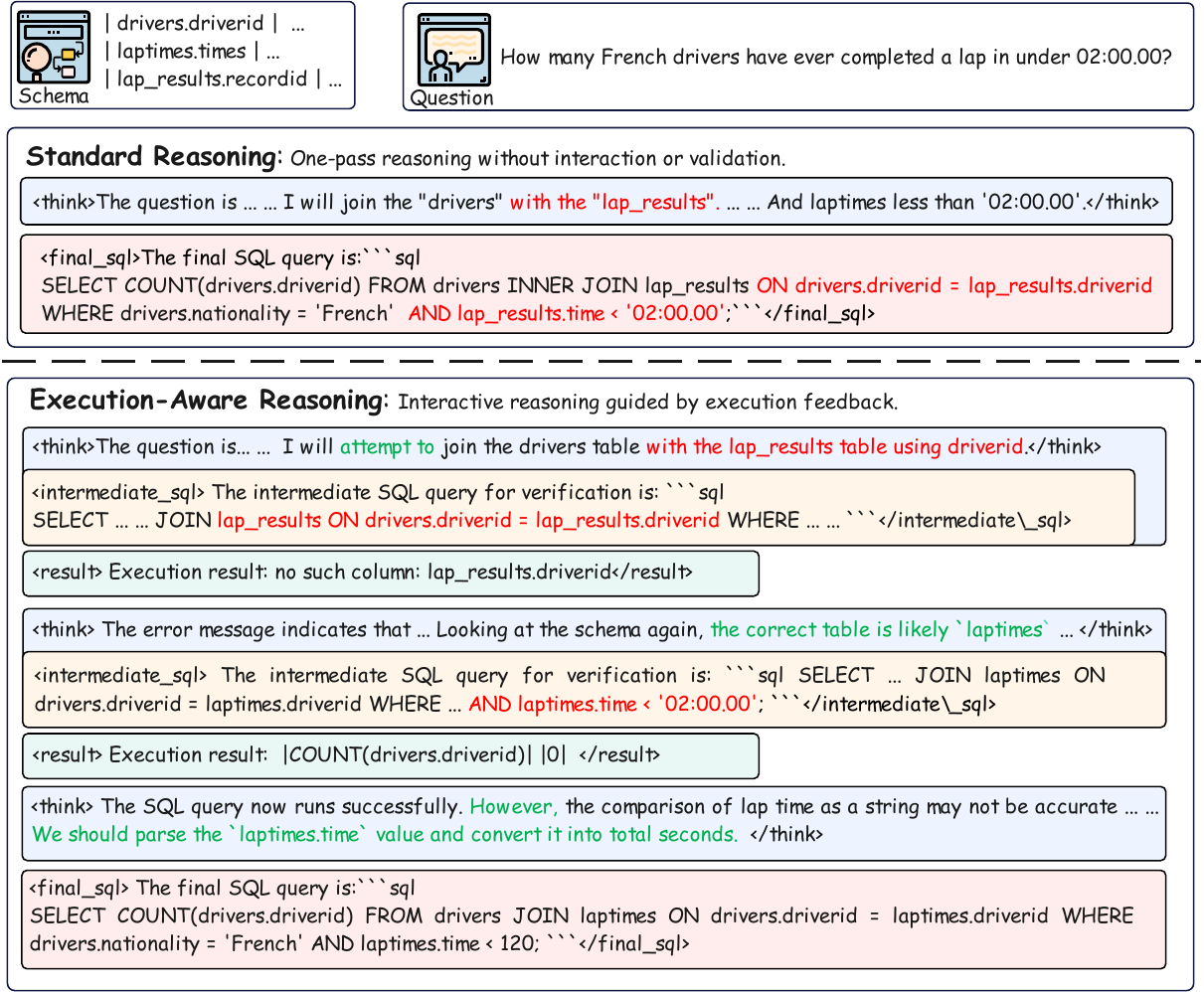} 
    \caption{Output comparison: execution-aware (ReEx-SQL) vs. standard reasoning (Baseline).}
    \label{fig:example_output_compare}
\end{figure}

This section compares the reasoning paths of ReEx-SQL and the Baseline to highlight the advantages of Execution-Aware Reasoning. As shown in Figure~\ref{fig:example_output_compare}, under the Standard Reasoning paradigm, the model relies solely on its internal knowledge and the given context, often resulting in reasoning errors and incorrect SQL generation. In contrast, Execution-Aware Reasoning allows the model to issue exploratory SQL queries to test its hypotheses and refine its reasoning based on execution feedback. For instance, in the example shown, the model receives an empty result from an exploratory query and correctly infers that the ``time'' column is an integer representing seconds rather than a string, leading to a corrected and accurate SQL query.

Beyond this case, ReEx-SQL demonstrates several generalizable behaviors driven by execution feedback. First, it can form and revise hypotheses through exploratory SQL based on observed results (e.g., Figure~\ref{example1}). Second, it is capable of detecting and correcting semantic errors, such as missing conditions, by interpreting empty result sets (e.g., Figures~\ref{example2} and \ref{example3}). Third, it can leverage execution error messages to adjust its reasoning, and when uncertain, it may re-run exploratory queries to validate its revisions (e.g., Figures~\ref{example4} and \ref{example5}). These examples collectively show that ReEx-SQL moves beyond static context-based reasoning by actively interacting with the environment, resulting in a more robust and adaptive SQL generation process.

\begin{figure}[H]
\begin{tcolorbox}[colback=yellow!1.5] 
\textbf{Question:} \\
\hspace*{0.2cm} The average unemployment ratio of 1995 and 1996, which one has higher percentage? \\
\textbf{Hit:} \\
\hspace*{0.2cm} a12 refers to unemploymant rate 1995; a13 refers to unemploymant rate 1996;\\
\rule{\linewidth}{0.4pt} 
\textbf{Standard Reasoning:} \\
\thinkstart  \\
The user's question asks for the average unemployment ratio of 1995 and 1996 and which one has a higher percentage.To answer this question, I need to look at the columns a12 (unemployment rate 1995) and a13 (unemployment rate 1996) in the district table.I need to calculate the average of these two columns to compare the percentages.I will start by selecting the columns a12 and a13 from the district table.Then, I will calculate the average of these two columns.\\
\thinkend \\
\finalstart\ The final SQL query is: ```sql \\
SELECT AVG(a12) AS avg\_unemployment\_1995, AVG(a13) AS avg\_unemployment\_1996 FROM district;''' \finalend \\
\rule{\linewidth}{0.4pt} 
\textbf{Execution-Aware Reasoning:} \\
\thinkstart  \\
Let's find the average unemployment ratio for both 1995 and 1996 from the `district' table. We need to calculate the average of `a12' (unemployment rate 1995) and `a13' (unemployment rate 1996). Then, we will compare these averages to determine which year has a higher percentage.\\
\thinkend \\
\sqlstart \\
The intermediate SQL query for verification is: ```sql\\
SELECT AVG(a12) AS avg\_1995, AVG(a13) AS avg\_1996 FROM district;'''\\
\sqlend \\
\resultstart \ |avg\_1995 | avg\_1996|  |3.0723684210526314 | 3.787012987012987| \resultend \\
\thinkstart \\
The intermediate SQL query correctly calculates the average unemployment rates for 1995 and 1996. The average unemployment rate for 1996 (3.787) is higher than for 1995 (3.072).\\
\thinkend \\
\finalstart\ The final SQL query is: ```sql \\
SELECT CASE WHEN AVG(a13) > AVG(a12) THEN `1996' ELSE `1995' END AS higher\_unemployment\_year FROM district;''' \finalend
\end{tcolorbox}
\caption{Reasoning output comparison for example 1.}
\label{example1}
\end{figure}

\begin{figure}[H]
\begin{tcolorbox}[colback=yellow!1.5] 
\textbf{Question:} \\
\hspace*{0.2cm} Which drivers who were born in 1971 and has the fastest lap time on the race? Give id and code of these drivers. \\
\rule{\linewidth}{0.4pt} 
\textbf{Standard Reasoning:} \\
\thinkstart  \\
The user's question asks for the drivers who were born in 1971 and had the fastest lap time on a race, along with their id and code.To answer this question, I need to focus on the ``dob'' column in the ``drivers'' table, which represents the date of birth of the drivers. I need to find drivers born in 1971, which means I need to look for rows where the year of ``dob'' is 1971. Additionally, I need to look at the ``fastestlaptime'' column in the ``results'' table, which represents the fastest lap time of a driver in a race. I need to join the ``drivers'' and ``results'' tables based on the ``driverid'' column to connect the information about the drivers' birth year and their fastest lap times.\\
\thinkend \\
\finalstart\\
The final SQL query is: ```sql \\
SELECT drivers.driverid, drivers.code FROM drivers INNER JOIN results ON drivers.driverid = results.driverid WHERE strftime('\%Y', drivers.dob) = '1971';'''\\
\finalend \\
\rule{\linewidth}{0.4pt} 
\textbf{Execution-Aware Reasoning:} \\
\thinkstart  \\
user is asking for drivers born in 1971 with the fastest lap time in a race. To solve this, we need to join the `drivers' table with the `results' table using the `driverid' column. We then filter for drivers born in 1971 by extracting the year from the `dob' column. The `results' table has a `fastestlaptime' column, which we will use to find the fastest driver. We'll begin with an initial query that selects all such drivers.\\
\thinkend \\
\sqlstart \\
The intermediate SQL query for verification is: ```sql\\
SELECT drivers.driverid, drivers.code, results.fastestlaptime FROM drivers JOIN results ON drivers.driverid = results.driverid WHERE STRFTIME('\%Y', drivers.dob) = '1971' ORDER BY results.fastestlaptime ASC LIMIT 1;'''\\
\sqlend \\
\resultstart \\
|driverid|code|fastestlaptime| \\ |14|COU|NULL| \\
\resultend \\
\thinkstart \\
The intermediate result shows that some `fastestlaptime' values are NULL. Including NULLs in sorting can lead to incorrect results, as NULL values do not represent valid lap times. Therefore, we must filter out rows where `fastestlaptime' is NULL before applying ordering and limits. We'll revise the SQL accordingly.\\
\thinkend \\
\finalstart\\
The final SQL query is: ```sql \\
SELECT drivers.driverid, drivers.code FROM drivers JOIN results ON drivers.driverid = results.driverid WHERE STRFTIME('\%Y', drivers.dob) = '1971' AND results.fastestlaptime IS NOT NULL ORDER BY results.fastestlaptime ASC LIMIT 1'''\\
\finalend 
\end{tcolorbox}
\caption{Reasoning output comparison for example 2.}
\label{example2}
\end{figure}

\begin{figure}[H]
\begin{tcolorbox}[colback=yellow!1.5] 
\textbf{Question:} \\
\hspace*{0.2cm} What was the release date for the set which card ``Evacuation'' in it? \\
\rule{\linewidth}{0.4pt} 
\textbf{Standard Reasoning:} \\
\thinkstart  \\
The user's question asks for the release date of the set that contains the card ``Evacuation''. To answer this question, I need to find the set code for ``Evacuation'' and then look up its release date. First, I need to locate the card ``Evacuation'' in the ``cards'' table by searching for its name.  Next, I need to join the ``cards'' table with the ``set\_translations'' table using the ``uuid'' column, as the set code is stored in ``set\_translations''.  Finally, I need to join the ``sets'' table with ``set\_translations'' using the ``code'' column to find the release date of the set.\\
\thinkend \\
\finalstart\\
The final SQL query is: ```sql \\
SELECT sets.releasedate FROM sets INNER JOIN set\_translations ON sets.code = set\_translations.setcode INNER JOIN cards ON set\_translations.id = cards.id WHERE cards.name = `Evacuation';'''\\
\finalend \\
\rule{\linewidth}{0.4pt} 
\textbf{Execution-Aware Reasoning:} \\
\thinkstart  \\
user is asking for the release date of the set that contains the card ``Evacuation''. To answer this question, we need to join the `cards' table with the `sets' table using the `uuid' column, which is common to both tables. The `cards' table contains the card name, and the `sets' table contains the release date of the set. By filtering the `cards' table for the card named ``Evacuation'', we can then extract the corresponding release date from the `sets' table.\\
\thinkend \\
\sqlstart \\
The intermediate SQL query for verification is: ```sql\\
SELECT sets.releasedate FROM cards JOIN sets ON cards.uuid = sets.mcmid WHERE cards.name = `Evacuation' '''\\
\sqlend \\
\resultstart\ |releaseDate| No data available in the database. \resultend \\
\thinkstart \\
The previous intermediate SQL query had a mistake in the join condition. The `sets' table's `mcmid' column does not match the `cards' table's `uuid' column. Instead, we should use the `sets' table's `code' column, which is referenced by the `cards' table's `setcode' column.  Let's correct the join condition and try again.\\
\thinkend \\
\sqlstart \\
The intermediate SQL query for verification is: ```sql\\
SELECT sets.releasedate FROM cards JOIN sets ON cards.setcode = sets.code WHERE cards.name = `Evacuation' '''\\
\sqlend \\
\resultstart \\
|releaseDate| |2007-07-13| |2003-07-28| |2001-04-11| Omit the sample below...... \\
\resultend \\
\thinkstart \\
The intermediate SQL query has been corrected to properly join the `cards' table with the `sets' table using the `setcode' column. The result shows that the release date for the set containing the card `Evacuation' is `2007-07-13'.\\
\thinkend \\
\finalstart\ The final SQL query is: ```sql \\
SELECT sets.releasedate FROM cards JOIN sets ON cards.setcode = sets.code WHERE cards.name = `Evacuation' ''' \finalend 
\end{tcolorbox}
\caption{Reasoning output comparison for example 3.}
\label{example3}
\end{figure}

\begin{figure}[H]
\begin{tcolorbox}[colback=yellow!1.5] 
\textbf{Question:} \\
\hspace*{0.2cm} How many posts were created on 21st July, 2010? \\
\textbf{Hit:} \\
\hspace*{0.2cm} created on 21st July, 2010 refers to creationdate BETWEEN `2010-07-21 00:00:00' and `2012-07-21 23:59:59';\\
\rule{\linewidth}{0.4pt} 
\textbf{Standard Reasoning:} \\
\thinkstart  \\
The user's question asks for the count of posts created on 21st July, 2010. To answer this question, I need to look at the creationdate column in the posts table.  I need to count the number of rows where the creationdate is between `2010-07-21 00:00:00' and `2010-07-21 23:59:59'. This will give me the number of posts created on that specific date.\\
\thinkend \\
\finalstart\\
The final SQL query is: ```sql \\
SELECT COUNT(*) FROM posts WHERE creationdate BETWEEN `2010-07-21 00:00:00' AND `2010-07-21 23:59:59' '''\\
\finalend \\
\rule{\linewidth}{0.4pt} 
\textbf{Execution-Aware Reasoning:} \\
\thinkstart  \\
user wants to find out how many posts were created on July 21, 2010. This requires querying the `posts' table, specifically the `creationdate' column, to count the number of posts that fall within the specified date range. The date range is from `2010-07-21 00:00:00' to `2010-07-21 23:59:59'. The SQL query will use the `COUNT' function to count the number of posts and the `WHERE' clause to filter the posts based on the `creationdate'. \\
\thinkend \\
\sqlstart \\
The intermediate SQL query for verification is: ```sql\\
SELECT COUNT(id) FROM posts WHERE creationdate BETWEEN `2010-07-21 00:00:00' AND `2010-07-21 23:59:59';'''\\
\sqlend \\
\resultstart\\ no such column: creationdate\\ \resultend \\
\thinkstart \\
The error indicates that the `creationdate' column does not exist in the `posts' table. The `creationdate' column is actually in the `posthistory' table, not the `posts' table. Therefore, the query needs to be adjusted to use the `posthistory' table instead.\\
\thinkend \\
\finalstart\\
The final SQL query is: ```sql \\
SELECT COUNT(id) FROM posthistory WHERE creationdate BETWEEN `2010-07-21 00:00:00' AND `2010-07-21 23:59:59' '''\\
\finalend 
\end{tcolorbox}
\caption{Reasoning output comparison for example 4.}
\label{example4}
\end{figure}

\begin{figure}[H]
\begin{tcolorbox}[colback=yellow!1.5] 
\textbf{Question:} \\
\hspace*{0.2cm} Please list the name of the cards in the set Coldsnap with the highest converted mana cost. \\
\rule{\linewidth}{0.4pt} 
\textbf{Standard Reasoning:} \\
\thinkstart  \\
The user's question asks to list the name of the cards in the set Coldsnap with the highest converted mana cost.  To answer this question, I need to find the cards in the set Coldsnap and look at their converted mana costs. I need to join the `sets' table with the `cards' table using the `code' column in `sets' and the `uuid' column in `cards'.  I need to find the card with the highest converted mana cost within the Coldsnap set.\\
\thinkend \\
\finalstart\\
The final SQL query is: ```sql \\
SELECT cards.name FROM cards INNER JOIN sets ON cards.uuid = sets.code WHERE sets.name = 'Coldsnap' ORDER BY cards.convertedmanacost DESC LIMIT 1;'''\\
\finalend \\
\rule{\linewidth}{0.4pt} 
\textbf{Execution-Aware Reasoning:} \\
\thinkstart  \\
user is asking for the names of cards in the set `Coldsnap' with the highest converted mana cost. To achieve this, we need to join the `cards' table with the `sets' table using the `uuid' column, which is common to both tables. The `sets' table will help us filter the cards by the set name `Coldsnap'. We will then sort the cards by their `convertedmanacost' in descending order to get the card with the highest converted mana cost at the top. Since we are only interested in the name of the card, we will select the `name' column from the `cards' table.\\
\thinkend \\
\sqlstart \\
The intermediate SQL query for verification is: ```sql\\
SELECT cards.name FROM cards JOIN sets ON cards.uuid = sets.uuid WHERE sets.name = 'Coldsnap' ORDER BY cards.convertedmanacost DESC LIMIT 1;'''\\
\sqlend \\
\resultstart\ no such column: sets.uuid \resultend \\
\thinkstart \\
The error indicates that we cannot directly join the `cards' table with the `sets' table using the `uuid' column because the `sets' table does not have a `uuid' column. Instead, we should use the `code' column from the `sets' table to match with the `code' column in the `cards' table. Let's correct the intermediate SQL query accordingly.\\
\thinkend \\
\sqlstart \\
The intermediate SQL query for verification is: ```sql\\
SELECT cards.name FROM cards JOIN sets ON cards.setcode = sets.code WHERE sets.name = 'Coldsnap' ORDER BY cards.convertedmanacost DESC LIMIT 1;'''\\
\sqlend \\
\resultstart\ |name|  |Sunscour| \resultend \\
\thinkstart \\
The intermediate SQL query has been corrected to join the `cards' table with the `sets' table using the `setcode' column, which is common to both tables. The query filters the cards by the set name `Coldsnap' and sorts them by their `convertedmanacost' in descending order. The result confirms that the card with the highest converted mana cost in the `Coldsnap' set is `Sunscour'.\\
\thinkend \\
\finalstart\\
The final SQL query is: ```sql \\
SELECT cards.name FROM cards JOIN sets ON cards.setcode = sets.code WHERE sets.name = 'Coldsnap' ORDER BY cards.convertedmanacost DESC LIMIT 1'''\\
\finalend 
\end{tcolorbox}
\caption{Reasoning output comparison for example 5.}
\label{example5}
\end{figure}

\end{document}